\pdfoutput=1
\documentclass{article}

\usepackage{arxiv}

\usepackage[utf8]{inputenc} 
\usepackage[T1]{fontenc}    
\usepackage{hyperref}       
\usepackage{url}            
\usepackage{booktabs}       
\usepackage{amsfonts}       
\usepackage{nicefrac}       
\usepackage{microtype}      
\usepackage{cleveref}       
\usepackage{graphicx}
\usepackage{multirow}
\usepackage{natbib}
\usepackage{doi}

\title{SeLeRoSa: Sentence-Level Romanian Satire Detection Dataset}

\date{}

\author{%
\begin{minipage}[t]{\textwidth}\centering
Răzvan-Alexandru Smădu$^{1}$ \quad
Andreea Iuga$^{1}$ \quad
Dumitru-Clementin Cercel$^{1,}$\thanks{Corresponding author.} \quad
Florin Pop$^{1,2}$\\[4pt]
{\normalfont
$^{1}$National University of Science and Technology POLITEHNICA Bucharest, Bucharest, Romania\\
$^{2}$National Institute for Research \& Development in Informatics -- ICI Bucharest, Bucharest, Romania\\[6pt]
\texttt{\{razvan.smadu,aiuga\}@stud.acs.upb.ro} \\
\texttt{dumitru.cercel@upb.ro} \;\; \texttt{florin.pop@cs.pub.ro}
}
\end{minipage}
}

\renewcommand{\headeright}{Smădu et al.}
\renewcommand{\undertitle}{}

\hypersetup{
pdftitle={SeLeRoSa: Sentence-Level Romanian Satire Detection Dataset},
pdfsubject={cs.CL},
pdfauthor={Răzvan-Alexandru Smădu, Andreea Iuga, Dumitru-Clementin Cercel, Florin Pop},
pdfkeywords={Satire Detection, Sentence-level Classification, Pretrained Language Models},
}

\begin{document}
\maketitle

\begin{abstract}
Satire, irony, and sarcasm are techniques typically used to express humor and critique, rather than deceive; however, they can occasionally be mistaken for factual reporting, akin to fake news. These techniques can be applied at a more granular level, allowing satirical information to be incorporated into news articles. In this paper, we introduce the first sentence-level dataset for Romanian satire detection for news articles, called SeLeRoSa. The dataset comprises 13,873 manually annotated sentences spanning various domains, including social issues, IT, science, and movies. With the rise and recent progress of large language models (LLMs) in the natural language processing literature, LLMs have demonstrated enhanced capabilities to tackle various tasks in zero-shot settings. We evaluate multiple baseline models based on LLMs in both zero-shot and fine-tuning settings, as well as baseline transformer-based models. Our findings reveal the current limitations of these models in the sentence-level satire detection task, paving the way for new research directions.
\end{abstract}


\section{Introduction}

Satire is a form of communication that employs techniques such as irony, sarcasm, and exaggeration to convey meanings opposite to those expressed on the surface, often with the intent to ridicule, critique, or expose social flaws. Satire, irony, and sarcasm have been studied together \citep{van-hee-etal-2016-monday,oprea-magdy-2020-isarcasm,10726230} in the literature along with humor detection \citep{10.1002/tesj.366} and sentiment analysis \citep{tay-etal-2018-reasoning,10488438,10.1145/3580496}. 

Satirical texts can deceive readers if they do not understand the subtle cues hidden in the text, giving a false sense of truth \citep{yang-etal-2017-satirical}. Satirical news is sometimes associated with fake news. However, fake news techniques are not always suitable for detecting satire \citep{rubin-etal-2016-fake}. For example, satirical news can be used to criticize and ridicule while being framed as legitimate articles \citep{10488438,yang-etal-2017-satirical}.

Most works focus on detecting satire at the article \citep{yang-etal-2017-satirical} or paragraph level \citep{yang-etal-2017-satirical}. However, recently, the desire to address sentence-level satire detection \citep{vaibhav-etal-2019-sentence} has increased. Sentence-level classification plays an important role, especially in fake news detection, where accurate information is often combined with false claims. For example, \citet{de-sarkar-etal-2018-attending} employed a hierarchical architecture that uses features to classify satirical news at the sentence and document levels. Their approach demonstrated that detecting key sentences can significantly enhance the document-level accuracy. \citet{10488438} utilized a classical architecture of Bidirectional Long-Short Term Memory \citep{DBLP:journals/neco/HochreiterS97} combined with a self-attention mechanism \citep{DBLP:conf/nips/VaswaniSPUJGKP17} that captures contextual information from short texts. They also showed that auxiliary features, such as scores related to sentiment analysis, punctuation, hyperbole, and effectiveness, can improve performance.
Most of the works addressed the English language \citep{khodak-etal-2018-large,oprea-magdy-2020-isarcasm,misra2023Sarcasm} for various domains, such as social media and news. However, existing works have also addressed satire detection in other languages, such as Arabic \citep{almazrua-etal-2022-sa7r}, Chinese \citep{yue-etal-2024-sarcnet}, French \citep{ionescu2021fresada}, Romanian \citep{rogoz-etal-2021-saroco}, and Spanish \citep{garcia-diaz2022compilation,alnajjar-hamalainen-2021-que}.

Recently, large language models (LLMs) have also been used in satirical news detection. \citet{ozturk-etal-2025-make} created a Turkish dataset for satire detection, along with a debiasing pipeline, utilizing LLMs to enhance the robustness and generalization of the models. \citet{DBLP:conf/aaai/YaoZLQ25} thoroughly analyzed LLM performance against various step-by-step following and prompting techniques, showing that sarcasm is not necessarily a sequential task, and reasoning may not always improve performance. Prompt engineering techniques for in-context learning were also evaluated by \citet{11036129} on various satirical datasets while proposing a prompting framework to improve overall performance.

To address the scarcity of resources in the Romanian language, we propose SeLeRoSa, the first \textbf{Se}ntence-\textbf{Le}vel \textbf{Ro}manian \textbf{Sa}tire detection dataset. The dataset contains 13,873 manually annotated sentences from satirical and factual news websites. We provide the anonymized dataset as is, as well as a pre-processed version. We evaluate several pre-trained LLMs, in zero-shot and fine-tuning settings, based on Gemma 2 and 3 \citep{gemmateam2025gemma3technicalreport}, Mistral \citep{jiang2023mistral7b}, and Llama 3.1 \citep{grattafiori2024llama3herdmodels}, as well as OpenAI's family of GPT-4o \citep{openai2024gpt4ocard}, GPT-4.1\footnote{\url{https://openai.com/index/gpt-4-1/}}, and the o4-mini reasoning models \citep{openai2024o3o4mini}.

In summary, our main contributions in this paper are as follows:
(1) we propose the first sentence-level dataset for Romanian satire detection from satirical news articles;
(2) we evaluate our dataset on several baselines based on pretrained language models;
(3) we make the code\footnote{\url{https://doi.org/10.5281/zenodo.15689793}} and dataset\footnote{\url{https://huggingface.co/datasets/unstpb-nlp/SeLeRoSa}} publicly available.

\section{Dataset}

\subsection{Data Collection}

The dataset was collected from satirical and non-satirical Romanian news outlets. The cut-off date is February 2018.
All data were scraped using Scrapy\footnote{\url{https://scrapy.org/}}. We first extracted 10,806 article titles, for which we performed manual filtering to remove news articles that contained inappropriate language, celebrity names, or references to sensitive or potentially controversial topics, including those related to religious institutions.
After this stage, we kept 3,692 titles. We then collected the content of the remaining news texts. From the resulting corpus, we removed markup tags, such as website-specific headers and HTML tags, and removed whitespaces. We retained diacritics if the text was written using them. Finally, we split each text into sentences and compiled a set of 36k sentences.

\subsection{Annotation Process}

Six native Romanian-speaking students annotated the final dataset over several years. The age of the students ranged from 20 to 25 years, with two male and four female annotators.
They received various parts of the dataset to annotate and were instructed to label the sentences as satirical, neutral, or uncertain.
The samples with uncertain labels were discarded from the final dataset. Ultimately, we compiled 13,873 sentences annotated by at least three annotators. The final label is determined by majority voting, i.e., the label that was common to at least two annotators.
This process yielded 8,179 regular and 5,694 satirical sentences.

\subsection{Inter-Annotator Agreement}

After obtaining the labeled dataset, we measure the inter-annotator agreement employing multiple metrics, as shown in Table \ref{tab:inter_annotator_agreement}. 
The Fleiss' Kappa (40.36\%) and Kendall's W (43.78\%) coefficients indicate moderate agreement among the annotators.
Furthermore, we compute the average of Cohen's Kappa for every pair of annotations, resulting in 40.53\% agreement between annotators, suggesting consistency among them. We also calculate the percentage agreement between the annotators as the proportion of labels on which the annotators agreed, which is 71.1\%. Finally, the human-level performance by comparing the labels assigned by the annotators with the majority voting label is 85.5\%. These results align with other existing works, showing the challenges and biases in annotating such data.

\begin{table}[!t]
\centering
\caption{Inter-annotator agreement statistics.}
\label{tab:inter_annotator_agreement}
\begin{tabular}{l|c}
\toprule
\multicolumn{1}{c|}{\textbf{Statistic}} & \textbf{Value} \\
\midrule
Fleiss' Kappa & 0.4036 \\
Kendall's W & 0.4378 \\
Average Cohen's Kappa & 0.4053 \\
Average Percentage Agreement & 71.1\% \\
Majority Voting Agreement & 85.5\% \\
\bottomrule
\end{tabular}
\end{table}

\subsection{Data Processing}

We provide two variants of our corpus: one that is only anonymized and one that is anonymized and pre-processed. 
Similar to \citet{butnaru-ionescu-2019-moroco} and \citet{rogoz-etal-2021-saroco}, we apply a processing step to reduce the potential leakage of satirical cues from the linguistic structures.

\textbf{Anonymization.} The anonymization stage implies identifying and removing the named entities from the text. We use Spacy's named entity recognition (NER) module\footnote{\url{https://spacy.io/models/ro\#ro_core_news_lg}} to identify persons, nationalities, religious or political groups, geopolitical entities, organizations, locations, and facilities in the text. We replace their occurrence with \texttt{<PERSON>}, \texttt{<NORP>}, \texttt{<GPE>}, \texttt{<ORG>}, \texttt{<LOC>}, and \texttt{<FAC>}, respectively. Additionally, we manually filter some frequently occurring entities that the NER model did not identify. We filter out URLs with \texttt{@URL} tag. In this light, our aim is also to reduce any potential unwanted biases related to personalities and institutions that could pose ethical issues (see also \S \ref{sec:ethics}). 

\textbf{Pre-processing.} The anonymized text is then pre-processed as follows. We convert the text to lowercase. Since some words use incorrect diacritics\footnote{According to the Romanian Association of Standardization (ASRO) in SR ISO/CEI 8859-16:2006, and the Romanian Academy in Orthographic, Orthoepic and Morphological Dictionary of the Romanian Language (DOOM)}, we replace ``ţ'' and ``ş'' (with cedilla) with ``ț'' and ``ș'' (with comma), respectively. Then, we use Spacy's tokenizer to split the text into tokens. We remove stop words, punctuation marks, and short tokens that have three characters or fewer. Finally, we lemmatize the text using Spacy's lemmatizer for the Romanian language. Throughout this processing stage, we maintain the tags used as placeholders unchanged in the anonymization stage.

This second version of the dataset can be used, for example, for text analysis (see \S \ref{sec:statistics}) and classical machine learning systems such as support vector machines \citep{DBLP:journals/ml/CortesV95}.

\subsection{Dataset Statistics}
\label{sec:statistics}

\begin{figure}[!ht]
\centering
\includegraphics[width=0.5\columnwidth,trim=0 20mm 0 10mm]{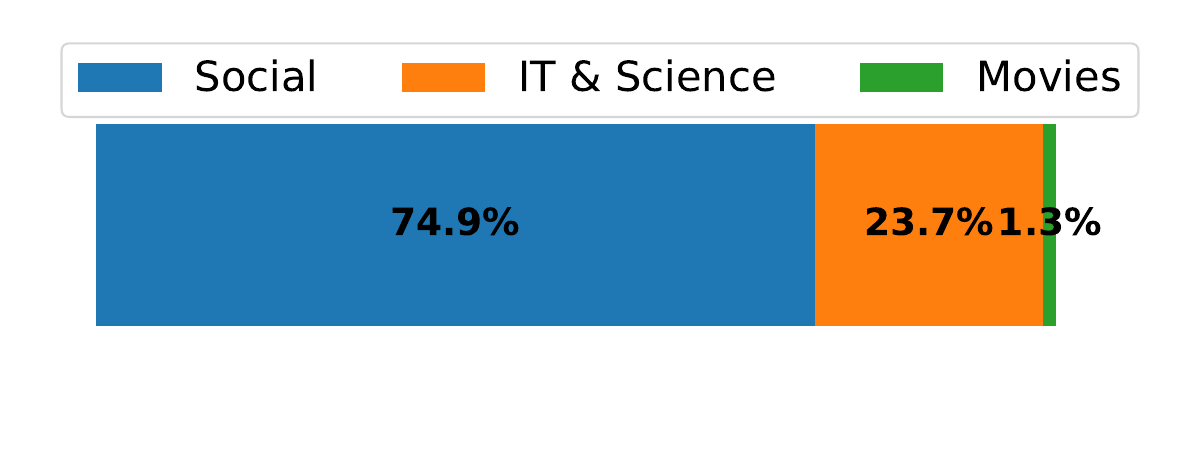}
\caption{Distribution of sentences across different domains in the dataset}
\label{fig:domain_dist}
\end{figure}

\begin{figure}[!ht]
    \centering
    \includegraphics[width=\columnwidth,trim=10mm 5mm 10mm 5mm]{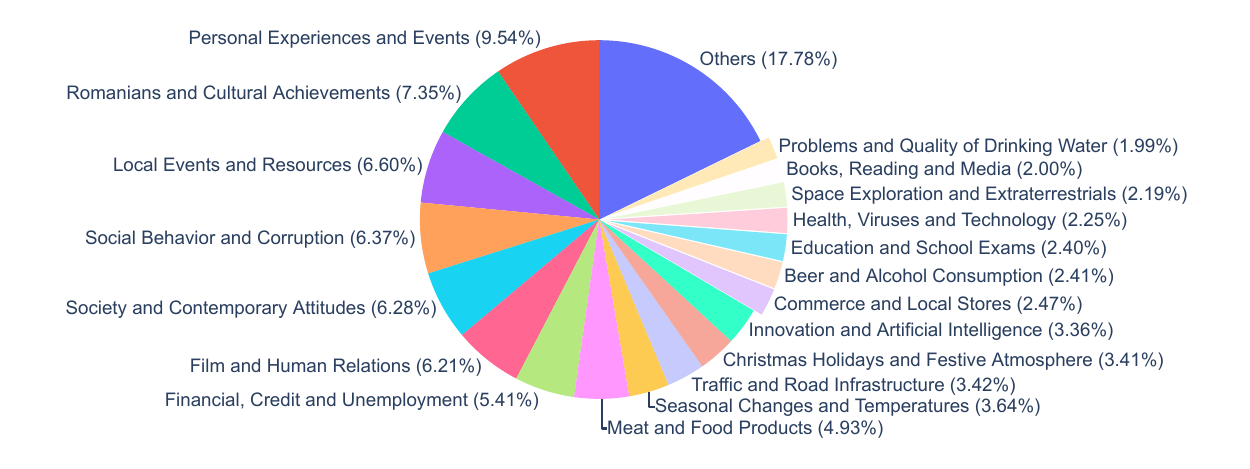}
    \caption{Topics distributions identified using BERTopic, on the entire dataset. We grouped topics with less than 2\% coverage under the \textit{Others} category.}
    \label{fig:topic_distr}
\end{figure}

The final dataset comprises sentences from three domains (see Figure \ref{fig:domain_dist}), with most of them representing the \textit{Social} domain, followed by the \textit{IT \& Science} and \textit{Movies} domains.


We extract and analyze the topics using BERTopic~\citep{grootendorst2022bertopic}.
BERTopic is a topic modeling technique that extracts coherent topics by leveraging the contextualized embeddings through the transformer architecture and a class-based TF-IDF approach. In our analysis, we employ the multilingual sentence embeddings\footnote{\url{https://huggingface.co/sentence-transformers/paraphrase-multilingual-MiniLM-L12-v2}} and set the minimum topic size to 50 samples. For the representation model, we employ the maximal marginal relevance and KeyBERT, in addition to cf-TF-IDF. We employ \texttt{gpt-4.1-mini-2025-04-14} to generate representative topic names. Using BERTopic, we identified 20 topics with at least 2\% coverage in the entire dataset, showcased in Figure \ref{fig:topic_distr}. The most representative topics are related to personal experiences, Romanian culture, corruption, and society.

We propose dividing the dataset into three sets: train, validation, and test. To avoid leaking contextual information about the satirical cues, the sentences from the same news article are grouped in the same split. Finally, the training set comprises 9,800 samples, while the validation and test sets each contain roughly 2,000 samples, as illustrated in Table \ref{tab:dataset_splits}. Most sentences are short, under 100 words, with a median length of approximately 19 words. Table \ref{tab:text_length_stats} shows detailed statistics for lengths in word and character count.

\section{Experiments and Results}

\begin{table}[!t]
    \centering
    \caption{Dataset splits and token distribution}
    \label{tab:dataset_splits}
    \setlength{\tabcolsep}{2pt}
    \begin{tabular}{l|cc|cc|cc}
        \toprule
        \multicolumn{1}{c|}{\multirow{2}{*}{\textbf{Split}}} &
        \multicolumn{2}{c|}{\textbf{Regular}} & 
        \multicolumn{2}{c|}{\textbf{Satirical}} &
        \multicolumn{2}{c}{\textbf{Total}} \\ 
            & \multicolumn{1}{c}{\textbf{Samples}}
            & \multicolumn{1}{c|}{\textbf{Tokens}}
            & \multicolumn{1}{c}{\textbf{Samples}}
            & \multicolumn{1}{c|}{\textbf{Tokens}}
            & \multicolumn{1}{c}{\textbf{Samples}}
            & \multicolumn{1}{c}{\textbf{Tokens}} \\
        \midrule
        Train      & 5,705 & 90,277 & 4,095 & 106,353 & 9,800 & 196,630 \\
        Validation & 1,240 & 19,979 & 760  & 20,058  & 2,000 & 40,037 \\
        Test       & 1,234 & 20,004 & 839  & 22,734  & 2,073 & 42,738 \\
        \midrule
        \textbf{Total} & 8,179 & 130,260 & 5,694 & 149,145 & 13,873 & 283,120 \\
        \bottomrule
    \end{tabular}
\end{table}

\begin{table}[!t]
    \centering
    \caption{Word and Character Length Statistics}
    \label{tab:text_length_stats}
    \setlength{\tabcolsep}{4pt}
    \begin{tabular}{l|cc|ccccc}
        \toprule
        \textbf{Metric} & \textbf{Mean} & \textbf{Std Dev} & \textbf{Min} & \textbf{Q1} & \textbf{Median} & \textbf{Q3} & \textbf{Max} \\
        \midrule
        Words & 20.1 & 10.1 & 1 & 13 & 19 & 26 & 96 \\
        Characters & 118.7 & 59.1 & 5 & 76 & 112 & 154 & 618 \\
        \bottomrule
    \end{tabular}
\end{table}

\subsection{Baselines}

We evaluate our dataset on several baselines. We finetune pre-trained language models such as the base Romanian BERT\footnote{\url{https://huggingface.co/dumitrescustefan/bert-base-romanian-cased-v1}} \citep{dumitrescu-etal-2020-birth} and the RoGPT2 large\footnote{\url{https://huggingface.co/readerbench/RoGPT2-large}} \citep{9643330}. For open-source LLMs, we employ the Gemma 3 family \citep{gemmateam2025gemma3technicalreport} of models, considering models with the parameter sizes of 1B, 4B, 12B, and 27B. These models were pre-trained on more than 140 languages, including Romanian. We also include LLMs specifically trained for the Romanian language \citep{masala-etal-2024-vorbesti} provided by the OpenLLM-Ro initiative\footnote{\url{https://openllm.ro}}. These models include RoMistral 7B, RoLlama 3.1 8B, and RoGemma 2 9B. In addition to those, we also include the Llama 3.1 8B model to compare the effects of Romanian language pre-training. For all locally run LLMs, we used their chat or instruction fine-tuned models when available. In the end, we evaluate the closed-source OpenAI models, including GPT-4o base and mini, GPT-4.1 nano, mini, and base, as well as the o4-mini reasoning model.

\subsection{Experimental Setup}

We run the experiments five times and report the mean and standard deviation. For evaluation metrics, we report accuracy and the F1-score, computed on the test set samples. For all experiments, we use the anonymized dataset only. The experiments were run on a single NVIDIA H100 80GB GPU.

\textbf{Pretrained Language Models.} For fine-tuning BERT, we use the Adam \citep{DBLP:conf/iclr/LoshchilovH19} optimizer with a weight decay of 0.01 and a learning rate of 2e-3 for 40 epochs. We set the batch size to 64. The model consisted of a single classification head on top of the frozen transformer model. On the other hand, RoGPT2 was treated similarly to an LLM, with the caveat that we disabled LoRA and quantization.

\textbf{LLM Fine-tuning.} We employ QLoRA \citep{10.5555/3666122.3666563} with 4-bit quantization. For LoRA, we set $R$=16, $\alpha$=32, dropout=0.05, and apply it to the attention layers. We use a learning rate with a linear scheduler, with the maximum value of 1e-4, and a 10\% warmup. We use the paged AdamW 8-bit optimizer. All models are trained for three epochs with a batch size of 32 and a maximum input length of 1024 tokens, which is sufficient to fit the prompt and the answer. 

\textbf{LLM Inference.} We use a default temperature of 0.8, top-k is set to 64, and top-p to 0.95. The maximum number of output tokens is set to 50, as the LLMs are expected to provide concise answers. The exception is for Gemma 3, which requires the temperature to be set to 1.0, and for the o4-mini, we use the default settings. For the reasoning model, we assign the reasoning effort to medium and set the maximum output tokens to 30,000. For inference-only models, we use a prompt that provides the format in which the model should give the output, together with the classes inserted in random order to avoid position bias \citep{10.5555/3666122.3668142}. For the fine-tuned models, we use a shorter prompt that does not provide instructions regarding the format, as this aspect is learned during fine-tuning.

\subsection{Results}

We present the results in Table \ref{tab:llm_results_f1_acc}. Most of the reported results are statistically significant with a p-value < 0.01 under the Student's t-test. Generally, we observe that the model size and whether the model was fine-tuned play a significant role in the performance. 

\begin{table}[!htbp]
\centering
\caption{Results of the models on the test set. The best results are highlighted in bold.}
\label{tab:llm_results_f1_acc}
\begin{tabular}{l|cc}
\toprule
\multicolumn{1}{c|}{\textbf{Model}}
    & \multicolumn{1}{c}{\textbf{Accuracy (\%)}}
    & \multicolumn{1}{c}{\textbf{F1-score (\%)}} \\
\midrule
RoBERT-cased FT & 76.58 $\pm$ 0.17 & 70.82 $\pm$ 0.44 \\
RoGPT2-large FT & 70.92 $\pm$ 0.40 & 65.77 $\pm$ 0.37 \\
\midrule
Gemma 3 1B & 42.69 $\pm$ 0.61 & 22.00 $\pm$ 1.81 \\
Gemma 3 4B & 44.50 $\pm$ 0.51 & 16.34 $\pm$ 1.15 \\
Gemma 3 12B & 56.37 $\pm$ 0.18 & 45.84 $\pm$ 0.49 \\
Gemma 3 27B & 50.71 $\pm$ 0.12 & 61.43 $\pm$ 0.05 \\
\midrule
RoMistral 7B & 45.21 $\pm$ 0.69 & 31.53 $\pm$ 1.35 \\
Llama 3.1 8B & 53.43 $\pm$ 0.55 & 39.57 $\pm$ 0.93 \\
RoLlama 3.1 8B & 54.54 $\pm$ 0.50 & 47.57 $\pm$ 0.70 \\
RoGemma 2 9B & 58.33 $\pm$ 0.26 & 55.10 $\pm$ 0.27 \\
\midrule
Gemma 3 1B FT & 53.96 $\pm$ 0.91 & 53.66 $\pm$ 0.91 \\
Gemma 3 4B FT & 71.02 $\pm$ 0.59 & 74.04 $\pm$ 0.38 \\
Gemma 3 12B FT & 76.22 $\pm$ 0.82 & 79.22 $\pm$ 0.72 \\
Gemma 3 27B FT & 74.61 $\pm$ 0.40 & 74.82 $\pm$ 1.93 \\
\midrule
RoMistral 7B FT & 74.12 $\pm$ 0.68 & 69.41 $\pm$ 0.74 \\
Llama 3.1 8B FT & 74.26 $\pm$ 0.48 & 71.19 $\pm$ 3.11 \\
RoLlama 3.1 8B FT & 73.39 $\pm$ 0.26 & 69.53 $\pm$ 0.46 \\
RoGemma 2 9B FT & \textbf{76.88 $\pm$ 0.66} & \textbf{80.72 $\pm$ 0.60} \\
\midrule
GPT-4o mini & 61.54 $\pm$ 0.18 & 64.81 $\pm$ 0.15 \\
GPT-4o & 62.99 $\pm$ 0.33 & 63.16 $\pm$ 5.29 \\
GPT-4.1 nano & 51.64 $\pm$ 0.46 & 47.20 $\pm$ 10.50 \\
GPT-4.1 mini & 61.64 $\pm$ 0.41 & 64.39 $\pm$ 0.36 \\
GPT-4.1 & 65.22 $\pm$ 0.24 & 67.60 $\pm$ 0.25 \\
\midrule
o4-mini (medium) & 72.53 $\pm$ 0.37 & 72.74 $\pm$ 3.40 \\
\bottomrule
\end{tabular}

\end{table}

The best-performing model is obtained by fine-tuning, yielding results with an accuracy over 70\% and an F1-score over 65\% (except for Gemma 3 1B). The highest scores are received by RoGemma 2 9B FT with 76.88\% accuracy and 80.72\% F1-score. We observe that generally, Romanian fine-tuned models perform marginally better in the zero-shot setting.

Evaluating the proprietary LLMs, we report that GPT-4.1 outperforms the GPT-4o family. We notice a minimal improvement in GPT-4o mini over GPT-4o, and even surpass it in the F1-score by less than 0.5\%. Considering the reasoning model o4-mini, it achieves fine-tuning-level performance, comparable to models with 4-10B parameters. This suggests the performance leap of reasoning models in the literature and their potential in future work, in assessing satire detection in news articles.

\subsection{Discussions}

Analyzing the predictions, we notice that, especially in the zero-shot setting, the models are more prone to Type I errors due to higher false positive rates (FPR). For the Gemma 3 models, we observe an FPR ranging from 32\% to 52\% of the number of samples. Fine-tuning the models reduces this to under 20\%. In all cases, the observed false negative rates (FNR) are less than 15\%. This suggests that models tend to overpredict the satirical class, and this can also be attributed to the system prompt, as we instruct the model to determine whether a sentence is satirical or not.

Investigating the performance over topics, we notice that less-representative topics, such as ``\textit{Sanctions and Legal Regulations}'' and ``\textit{Problems and Quality of Drinking Water}'', are easier for the models to classify, with an F1-score of over 70\%. In contrast, larger topics such as ``\textit{Financial, Credit and Unemployment}'', ``\textit{Smoking and Quitting Methods}'', and ``\textit{Books, Reading and Media}'' are more challenging, where the average F1-score is 56.8\%, 54.7\%, and 54.4\%, respectively. Additionally, we observe consistent overall scores across topics for the fine-tuned models. 

\section{Conclusion}

In this work, we proposed SeLeRoSa, the first sentence-level Romanian satire detection dataset. It was manually annotated, covering 8,179 regular sentences and 5,694 satirical sentences. It contains several domains and spans multiple topics. We presented several baseline models evaluated on this dataset, ranging from small pre-trained language models to open-source and closed-source language models in zero-shot and fine-tuning settings. Our experiments showed that models tend to overpredict the satirical class, whereas the latest advancements in reasoning models may achieve performances similar to those of fine-tuned models. We hope this dataset opens new future directions in sentence-level satire detection and advancements in the low-resource Romanian language.

\section*{Ethical Considerations}
\label{sec:ethics}

The original copyright of the satirical sentences used in our work remains with the original authors of the news articles, as per existing copyright laws. We release the dataset for academic purposes, under the CC BY-NC-SA 4.0 license.
Additionally, we anonymized the entities to minimize potential bias and discrimination against the involved entities. Any statements from the dataset were made solely by the authors of the original news articles, and we do not share their views nor promote any potentially harmful content. To the best of our ability, we tried to filter out those examples through manual inspection and automatic tools.

\section*{Acknowledgments}
This work is supported by the project \textit{Romanian Hub for Artificial Intelligence - HRIA}, Smart Growth, Digitization and Financial Instruments Program, 2021-2027, MySMIS no. 334906, and by the National University of Science and Technology POLITEHNICA Bucharest through the PubArt program. 
We also thank Daiana Brabie, Dana Ciobanu, Liviu Andrei Mureşanu, and Sandra-Mihaela Țuca for their contribution in annotating the datasets, and Brînduşa Ioana Niculae for the help in curating the data.

\section*{GenAI Usage Disclosure}

In this work, we disclose that we did not include GenAI content. We used large language models as evaluation baselines for our approach, just as we would for any other neural network-based approach. Additionally, the manuscript was written solely by the authors. Most of the data collection and annotation stages were completed before 2023, before GenAI technologies, such as ChatGPT, became widely popular for automating workflows. The code we release is also not generated by AI.

\bibliographystyle{unsrtnat}
\bibliography{bibliography}

\end{document}